\documentclass[letterpaper, 10 pt, conference]{ieeeconf}  
\IEEEoverridecommandlockouts                              

\usepackage[backend=biber,style=numeric,sorting=none]{biblatex} 
\bibliography{library} 

\usepackage{graphicx}
\usepackage{subfigure}
\usepackage{float}

\usepackage{color}
\usepackage{makecell}
\usepackage{CJK}
\newcommand{\tabincell}[2]{\begin{tabular}{@{}#1@{}}#2\end{tabular}}

\usepackage{algorithm}
\usepackage[noend]{algpseudocode}
\usepackage{amssymb}
\usepackage{amsmath}
\usepackage{dsfont}

\usepackage{amsthm}
\usepackage{csquotes}
\usepackage{hyperref}
\usepackage{float}
\usepackage{pdfpages}
\usepackage{fancyhdr}
\upshape
\makeatletter

\newcommand{\Rmnum}[1]{\expandafter\@slowromancap\romannumeral #1@}
\makeatother

\title{\LARGE \bf
AB-Mapper: Attention and BicNet Based Multi-agent Path Finding for Dynamic Crowded Environment
}

\author{Huifeng Guan$^{1*}$, Yuan Gao$^{2*}$, Min Zhao$^{2}$, Yong Yang$^{4}$, Fuqin Deng$^{1,2,4,\dagger}$, Tin Lun Lam$^{2,3,\dagger}$

\thanks{$^{*}$ Authors contributed equally, ranked alphabetically.
        $^{1}$ School of Intelligent Manufacturing, the Wuyi University, Jiangmen, China.
        $^{2}$ Shenzhen Institute of Artificial Intelligence and Robotics for Society, Shenzhen, China.
        $^{3}$ School of Science and Engineering, the Chinese University of Hong Kong, Shenzhen, China.
        $^{4}$ 3irobotix Co., Ltd, Shenzhen, China.
        $^{\dagger}$ Corresponding authors are Tin Lun Lam tllam@cuhk.edu.cn and Fuqin Deng fuqindengteam@163.com.
}
\thanks{This work was supported in part by the National Key R\&D
Program of China (2020YFB1313300), the funding (AC01202101025,
AC01202101026) from the Shenzhen Institute of Artificial Intelligence
and Robotics for Society, the special projects in key fields of Guangdong Provincial Department of Education (2019KZDZX1025), the Basic and Applied Basic Research Foundation of Guangdong under Grant (2019A1515111119, 2021A1515010926), Innovative Program for Graduate Education (503170060259) from the Wuyi University, and Shenzhen Peacock Plan of Shenzhen Science and Technology Program (KQTD2016113010470345).}
}

\begin{document}
\maketitle
\thispagestyle{empty}
\pagestyle{empty}

\begin{abstract}
Multi-agent path finding in dynamic crowded environments is of great academic and practical value for multi-robot systems in the real world. 
To improve the effectiveness and efficiency of communication and learning process during path planning in dynamic crowded environments, we introduce an algorithm called Attention and BicNet based Multi-agent path planning with effective reinforcement (AB-Mapper) under the actor-critic reinforcement learning framework. In this framework, on the one hand, we utilize the BicNet with communication function in the actor-network to achieve intra team coordination. On the other hand, we propose a centralized critic network that can selectively allocate attention weights to surrounding agents. This attention mechanism allows an individual agent to automatically learn a better evaluation of actions by also considering the behaviours of its surrounding agents. 
Compared with the state-of-the-art method Mapper, our AB-Mapper is more effective (85.86\% vs. 81.56\% in terms of success rate) in solving the general path finding problems with dynamic obstacles. In addition, in crowded scenarios, our method outperforms the Mapper method by a large margin, reaching a stunning gap of more than 40\% for each experiment. 
\end{abstract}

\section{Introduction}
The Multi-agent path finding (MAPF) problems refers to solving the path planning problems for multiple agents. These agents plan to reach their goals from their starting positions while avoiding conflicts among themselves and with other agents that exist in environment~\cite{lejeune2021survey}.
In crowded environments, the MAPF problems are a challenging problem and have attracted the attention of many research groups in recent years~\cite{vsvancara2019online,stern2019multi}.

In general, there are two classes of algorithms commonly used for solving MAPF problems. One class of them is the A*-based path planning method.
In this class, the collision-based search (CBS) is one of the mainstream methods~\cite{li2019symmetry}. The CBS could be summarized into two main steps. For the first step, each agent plans its path based on A*. In this step, each agent ignores the presence of other agents; For the second step, the central planner detects if there is a collision. Then the
collision node needs to store all scheduling allocation schemes for the constrained actions.
Therefore,
the CBS can efficiently solve the MAPF problems in a static 
environment~\cite{felner2017search}. If the number of conflicts between agents is relatively small, the CBS does not need to spend too much computational power to draw out scheduling allocation schemes. However, when the number of actions increases in dimension, it leads to an exponential growth in the length of the set of scheduling allocation parties with respect to the number of agents~\cite{felner2017search}, especially in a crowded environment.
This challenge leads to an increasingly long time and it may exhaust the available memory and computational resources~\cite{sharon2015conflict}. As a consequence, the CBS has weak generalization capabilities and lacks the ability to handle dynamic obstacles, therefore it is not suitable for MAPF in dynamic crowded environments.

\begin{figure}[t]
    \centering
    \includegraphics[width=0.5\textwidth]{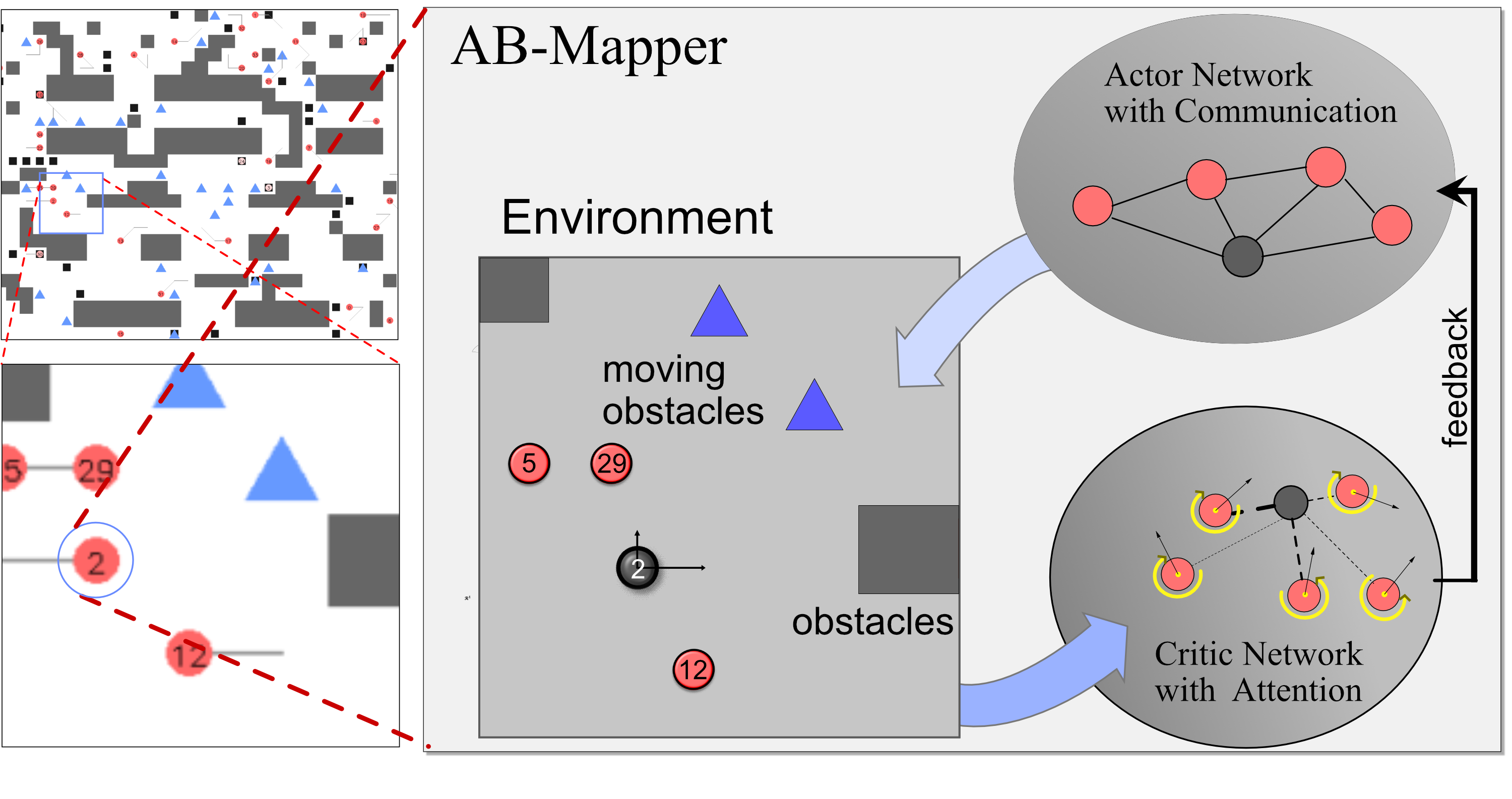}
    \caption{Main ideas of AB-Mapper algorithm. The actor network of AB-Mapper utilizes BicNet to emphasize the importance of communication between agents, whilst the critic network uses the attention mechanism to evaluate the performance of the actor network by selectively paying attention to important agents in the environment.}
    \label{fig:ab_Mapper}
    \vspace{-7mm}
\end{figure}
The second mainstream approach is deep reinforcement learning based algorithms. Unlike the above-mentioned algorithms, they have been gradually adopted in recent years to solve MAPF problems with strong environment generalization and adaptative capabilities~\cite{hasan2020defensive}. In general, deep reinforcement learning algorithms can be broadly classified into two categories based on the training 
framework: distributed training distributed execution framework (DTDE) and centralized training distributed execution framework (CTDE). In this paper, we develop
a deep reinforcement learning algorithm with CTDE architecture to consider the evaluation of joint strategies for path planning in multi-robot systems.
To improve the planning of multi-intelligent systems in dynamic crowded environments under the actor-critic reinforcement learning framework, we introduce a BicNet network into the actor network, allowing the agent to pass information to other agents. Furthermore, recognizing the importance of evaluating the states and actions of surrounding agents, we introduce an attention mechanism into the critic network. As a whole, our algorithm encodes the features extracted by the convolutional neural network (CNN), together with the planned actions
from the actor network, allowing more structured information to be fed to the critic network. At the same time, the critic network with attention mechanism evaluates the actor network not only by considering the state-action information and the $Q$ value of the agent, but also by selectively paying attention to the state-action information of a limited number of other agents in the surrounding area to participate in the decision. This novel design greatly improves the intelligence of the decision (see Fig.~\ref{fig:ab_Mapper}).
Our major contributions are:
\begin{itemize}
    \item We discuss the importance of integrating BicNet in actor network, so that the agents can plan their actions with the state information of other agents. We show that this enables communication among agents, leading to speeding up the convergence of the algorithm.
    \item We introduce the attention mechanism into our AB-Mapper's critic network. Each agent uses this network to selectively pay attention to the actions and states of other agents within a limited number of surrounding regions, enabling the agent to know more precisely which surrounding agents should be more beneficial to it in the path planning process. This leads to the stability and convergence of the algorithm for MAPF in a dynamic crowded environment.
\end{itemize}

This paper is organized as follows: 
In Section \Rmnum{2}, we review related works on MAPF problems. In Section \Rmnum{3}, we continue with briefing the technical background of our proposed AB-Mapper method and introducing its working principles, including the BicNet network, the attention mechanism, and relevant details. In Section \Rmnum{4}, we show setups of our experiments and analysis of our results. Finally, we sum up our study in Section \Rmnum{5}.

\section{Related works}
In this section, we first review previous works related to deep reinforcement learning based methods applied in MAPF problems. Then we continue discussion regarding agent communication problems in MARL, which is highly related to the BicNet used in our study. After that, we briefly go through the development of the attention mechanisms used in our critic networks.

\subsection{MAPF based on Deep Reinforcement Learning} 
In recent years, solving MAPF problems by deep reinforcement learning methods has attracted the attention of researchers, and the algorithms can be classified into non-communicative and communicative methods according to whether the information is exchanged between the agents.
In the communication-free path planning approach, it is common to generate prior knowledge using expert strategies and then train the neural network by minimizing the loss of a given state-action. For example, Wang et al.~\cite{wang2020mobile} introduced a globally guided reinforcement learning method to obtain a single optimal path, followed by a fixed globally guided case based on DDQN to plan out the action by using the information about the surrounding environment. Chen et al.~\cite{chen2017decentralized} proposed a method of supervised training for the value network through a set of trajectories generated by the baseline strategy OCRA when the expected target time of learning to code is given. Similarly, Riviere et al.~\cite{riviere2020glas} 
presented an algorithm for training neural networks by minimizing the loss of output actions for a given observation after obtaining batches of observation-action pairs from an expert presentation dataset.
Among these methods, the most classical one is the PRIMAL~\cite{sartoretti2019primal} method, which is based on using A* to plan the local trajectories of all agents in the operation area and fusing the basic path conflict constraint matrix into the state set. This approach uses the ODrM* algorithm to plan the initial paths and a manually defined expert system to repair the initial paths on the one hand, and incorporates the processed paths into the actor-critic framework for training on the other hand. However, the information between each agent in PRIMAL is not available for interaction and does not consider non-cooperative dynamic obstacles nor temporal information. Therefore, imitation learning takes longer time to train intensively.

Later on, Liu et al.~\cite{liu2020mapper} proposed the Mapper, which is the baseline method of our study, for MAPF under the DTDE architecture. In this method, each agent models the behavior of dynamic obstacles based on the image-based representation and then inputs local observations into the respective actor and critic networks for learning. In addition, Mapper designs a dynamically adjustable evolution probability by normalizing the reward values obtained by all agents after a fixed number of iterations. Then Mapper samples the network parameters from a distribution over the agents that received a relatively large reward. This mechanism enables Mapper to perform path planning more efficiently and effectively than PRIMAL in a dynamic environment.

Most of the above communication-free methods are based on DTDE framework. In this framework,each agent plans actions based only on its sequence of observations and its policy~\cite{liu2020mapper}, which is feasible in a non-crowded environment where each agent's decision does not have to consider other agents because the combination of the optimal individual actions is the optimal joint action, and there is no need to consider the communication between the agents, which makes the distributed approach highly efficient~\cite{chen2017decentralized}. However, the interaction between the agent and the dynamic crowded environment makes the agent have to overcome the problem of low stability and poor robustness of the planned strategies due to the non-stationarity of the environment~\cite{omidshafiei2017deep}. In this paper, we focus CTDE framework which allows each agent to learn and construct its behavioral value function after aggregating the state and action information of other agents ~\cite{son2019qtran}. Therefore, the agent can learn whether the actions of other agents are beneficial or harmful to itself, which helps to mitigate the issue of non-stationarity during training in the dynamic crowded environment ~\cite{wadhwania2019policy}.
\subsection{Communication in MARL}
Like the group foraging behavior in nature and the coordination of human society, each participant in the group has a one-sided understanding of the environment. It is inevitable to learn coordinated actions to complete the task efficiently. In the MARL environment, information exchange plays a vital role in the coordinated behavior among agents. Therefore, establishing an efficient communication protocol for agents has become the focus of researchers in recent years.
For example, Sukhbaatar et al. proposed CommNet based on continuous communication for the task of fully cooperative mode in a fully observable environment. CommNet can fuse and transfer information, which indirectly considers the global state output strategy~\cite{sukhbaatar2016learning}. Singh et al. designed the IC3Net for the task of competitive mode, which controlled a communication-related binary gating function through the LSTM network gating mechanism and prevented the communication of multiple agents in a competitive relationship~\cite{singh2018learning}. To better deal with the fully cooperative, partially observable multi-agent sequential decision problem, Foerster et al. developed the DIAL, in which, communication information is passed through a discrete regularization unit (DRU) between the output of one network of agents and the input of another network of agents. The DRU had a bidirectional information transfer function. At the optimization stage, the gradient could be returned along the channel, thus allowing the end-to-end backpropagation across the agents ~\cite{foerster2016learning}.
\subsection{Attention Mechanisms}
Many researchers have tried to apply the attention mechanism in MAPF. For example, Jiang et al.~\cite{jiang2018graph}, and Ma et al.~\cite{ma2021distributed} modeled the relationship within a multi-agent environment using a graph, where the nodes of the graph are the agents and the encoding of the local observations of the agents are the features of the nodes. This work employed multi-headed attention as a graph convolution kernel to extract the relational representations among the agents and convolve the potential features from the neighboring nodes. Zhang et al.~\cite{zhang2020robot} used the multi-headed attention mechanism to calculate the weight of the current agent's interaction with other agents. Zhou et al.~\cite{zhou2021robot} introduced the attention mechanism into a graph attention network that modeled different traffic participants so that both the agent-human interaction and the human-human interaction model could be described by graphs using the same graph convolution operation. Chen et al.~\cite{chen2020robot} introduced a graph convolutional network (GCN) based on human gaze data. This network can predict human attention to different subjects in the crowd, and then the learned attention weights are integrated into the adjacency matrix of the GCN to manage and aggregate features for estimating robot-crowd states.

In contrast to the multi-headed attention mechanism approach based on graph neural networks, some algorithms do not model the agents as graphs. For example, Rosbach et al.~\cite{rosbach2020planning} applied the attention mechanism to inverse reinforcement learning for predicting the reward function over an extended planning horizon. Shah et al.~\cite{shah2018follownet} proposed a novel linguistic instruction attention mechanism, where the attention mechanism scored the tokens of the input visual and instruction information. This method finally used a softmax function to normalize the relative importance of each token corresponding to the current instruction. Chen et al. ~\cite{chen2019crowd} proposed to use an attention mechanism to learn the collective importance of neighboring humans' behavior for their future states, capturing human-human interactions in dense crowds.
This mechanism can indirectly influence the predictive ability of the crowd dynamics and reduce collision rates while navigating the ways for the agents.
\section{Proposed Approach For Path Planning}
\subsection{Preliminaries}
In reality, due to the asymmetry of information, human beings have to make decisions based on part of the information. Similarly, in this paper, we model the interaction process between agent and environment as partially observed Markov decision processes (POMDPS) tuple, ($\mathcal{S}$, $\mathcal{A}$, $P$, $R$, $\mathbf{o}$, $\vartheta,\gamma$). $\mathcal{S}$ represents the state space, and $\mathcal{A}$ represents the action space. $P: S \times A \times S\to[0,1]$ denotes the transition probability.  $R: S \times A \to \mathbb{R}$ is a reward function. $\mathbf{o}$ represents local observation. $\vartheta$ denotes conditional local observation probability and $\gamma\in[0,1] $ is a discount factor~\cite{choi2020fast}.

Since the Mapper method under the DTDE framework cannot efficiently circulate information between the agents, nor can it plan a coordinated policy in a dynamic crowded environment, resulting in slow or even difficult convergence, we present our AB-Mapper method under the CTDE framework, in which agents use centralized information for offline training, but execute online in a decentralized manner~\cite{lyu2021contrasting}. This learning method is very suitable for the actor-critic structure, which has gradually become popular in the field of MARL~\cite{wen2020smix}. The actor-critic method estimates the expected return $Q$ of a given state and action by learning a function $Q_{j}\left(\mathbf{o_{j}}, a_{j}\right)$ for the agent $j$ and learns through off-policy temporary difference learning by minimizing the region loss.
In AB-Mapper, the loss function of actor network is defined as
\begin{equation}
\begin{array}{r}
\mathcal{L}_{A}=-\sum_{j=1}^{n}\left(\log \pi_{\theta}\left(a_{j} \mid \mathbf{o_{j}}\right) * y\right) 
\end{array}
\label{equ:actor}
\end{equation}
\begin{equation}
    \\y=r_{j}+\gamma Q_{j}\left(\mathbf{o_{j}^{\prime}}, a_{j}^{\prime}\right)
    \label{equ:y}
\end{equation}
, where the $\mathbf{o_{j}^{\prime}}$ and $a_{j}^{\prime}$ are the local observation and action of the agent $j$ at the next moment respectively. The critic network's loss function is
\begin{equation}
\begin{array}{r}
\mathcal{L}_{C}=\sum_{j=1}^{n}\left(Q_{j}\left(\mathbf{o_{j}}, a_{j}\right)-y\right)^{2}.
\end{array}\label{equ:critic}
\end{equation}
\subsection{Method}
The flow of the AB-Mapper algorithm is mainly composed of three steps: first, a CNN is used to describe the state and output the features, then the features are inputted into BicNet to plan the policy, and finally, the features and policy are fed into the critic network for joint optimization. 

\begin{figure}[t]
\vspace*{8pt} 
\centering
\includegraphics[width=0.45\textwidth]{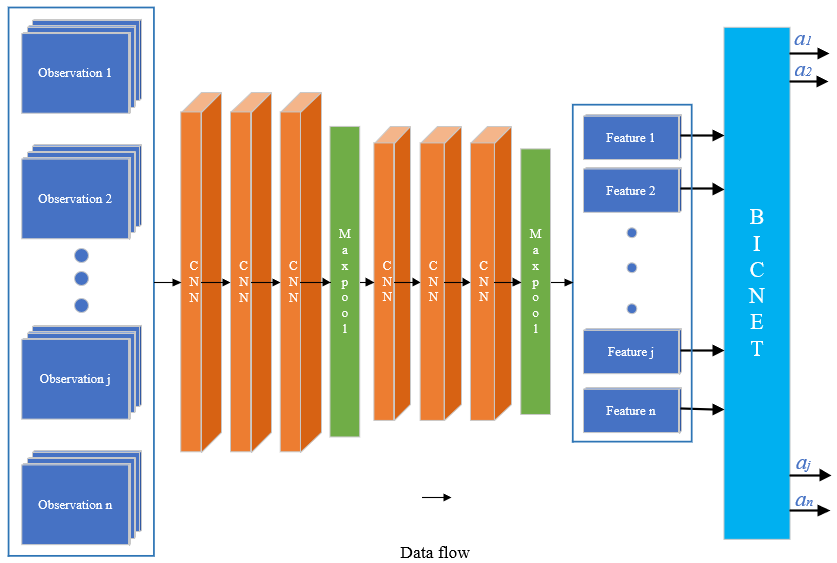}
\caption{ AB-Mapper's actor network. We use CNN to extract features from the local observations collected from all agents at one time and enter them into BicNet to plan coordinated actions.}\label{fig:actor_v1}
\vspace{-6.5mm}
\end{figure}

Different from Mapper's description of local observations of an individual agent using a CNN respectively, we first collect the local observations of all the agents in the actor network of AB-Mapper method as shown in 
in Fig.~\ref{fig:actor_v1}.
Then, we develop a 6-layer CNN to output all features used for BicNet to plan policy. The BicNet has the structure of a bidirectional recurrent neural network~\cite{peng2017multiagent}.
Such the network has forward and backward hidden layers so that the output node at each timestep contains complete past and future contextual information at the current moment in the input sequence. At each step of sequential decision-making, each agent can maintain its own internal state and share information with its collaborators, thus coordinating the agents' actions.
We choose BicNet as part of the actor network because the parameter space of BicNet communication is independent of the number of agents. Also, unlike CommNet and other communication networks that require information aggregation, the BicNet network can efficiently transfer information by simply inputting the state information of the time series.
It is suitable for the action decisions of the agents in path planning and can be perfectly integrated with our experimental setup for path planning.

\begin{figure}[t]
\vspace*{8pt} 
\centering
\includegraphics[width=0.45\textwidth,height=0.31\textheight]{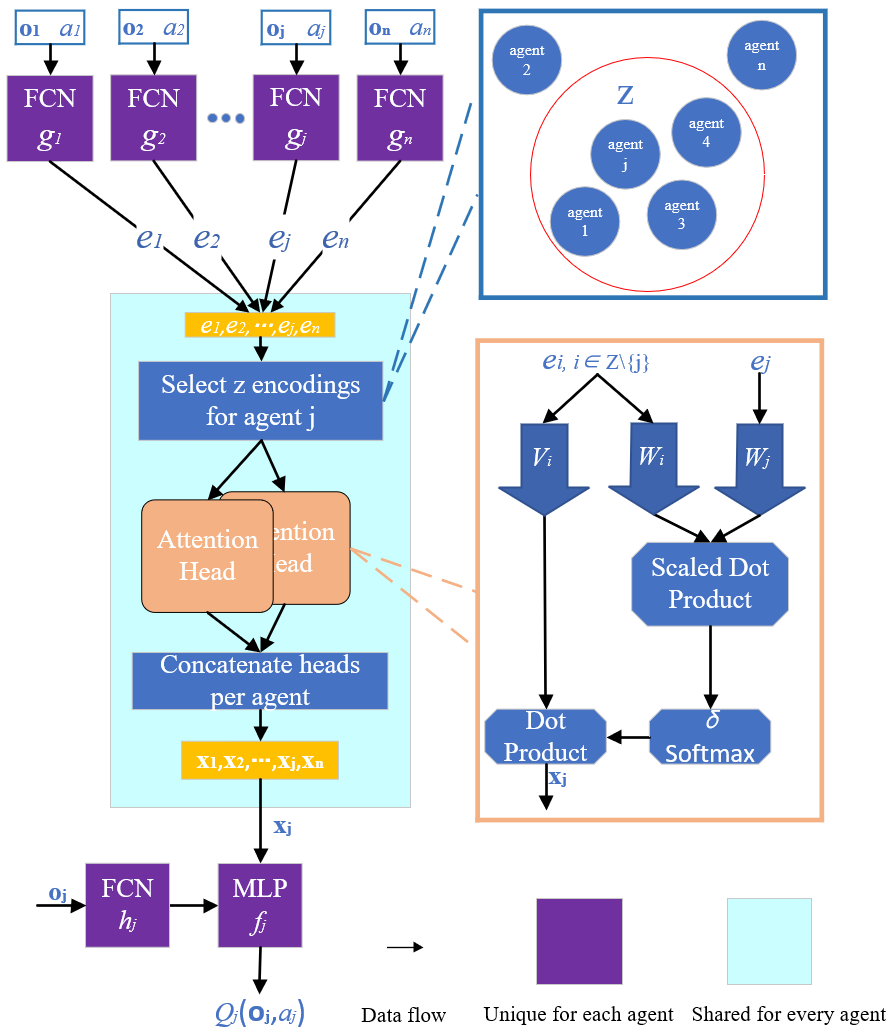}
\caption{AB-Mapper's critic network. Calculating $Q_j (\mathbf{o}_j,a_j )$ with attention for agent $j$. Agent $j$ encodes its local observations and action, receives a weighted sum of z agent encodings, sends it to the attention mechanism. The $W_j$ transmutes $e_j$ into a query $\mathbf{q}_j$ and $W_i$ transmutes $e_i$ into a key $\mathbf{k}_i$. The $V$ transmutes $e_i$ into a value $\mathbf{v}_i$.}
\label{fig:critic_v1}
\vspace{-5mm}
\end{figure}

In Mapper, each agent has an independent critic network, but it lacks interaction of information between critic networks, making it impossible for a single agent to know which agent's state action information 
may have an impact on itself.
Applying attention mechanism to critic network is an effective method to solve the above problem. The main idea of the attention mechanism is to calculate the attention weight through three elements: query, key, and value~\cite{vaswani2017attention}.
It is a mainstream practice to influence the prediction of $Q$ values by calculating attention weights, as used in the multi-agent environments of ~\cite{iqbal2019actor} and~\cite{parnika2021attention}. In these two works, the number of agents is small, and the state-action information of all agents is involved in calculating the $Q$ value of one agent by a centralized critic network with attention. However, in an experimental setting with a large environmental dimension and many agents, it is impractical for each agent to pay attention to all the remaining agents because some agents are so far apart that their local states and actions have weak effects current agent. Therefore, there is no need to assign attention to agents far away.
This attention mechanism can save many computing resources for nearby agents. Recently, Li et al.~\cite{li2020graph} proposed a method to assign the attention weight of an agent to all agents in the neighborhood with radius R. Similarly, in our AB-Mapper method, 
the critic network learns agents' judgments by selectively focusing on the state and action information of a limited number of other agents around, as shown in Fig.~\ref{fig:critic_v1}. This centralized learning of judgments allows each agent to learn the state and action of the surrounding agents that affect the current agent, redistributing focuses to the agents that should be attended to.
Therefore, in this paper, in order to adapt to the path planning task of multi-agent in a crowded environment, we impose a restriction on the number of attended agents on the critic network with an attention mechanism, as detailed below. We encode the state-action information ($\mathbf{o_1}$, $a_1$, $\mathbf{o}_2$, $a_2$, ..., $\mathbf{o}_j$, $a_j$, $\mathbf{o}_n$, $a_n$) of all agents through their respective encoders ($g_1$, $g_2$, ..., $g_j$, $g_n$) and input it to the critic network. Each encoder is a one-layer fully connected network (FCN), specified as 
\begin{equation}
    \mathbf{e}_j=g_j (\mathbf{o}_j,a_j ).
\label{equ:e_j}
\end{equation}

At every timestep of path planning for agent $j$, we input the encoded information of $\mathbf{o}_j$ by another FCN $h_j(\cdot)$, and $\mathbf{x}_j$ into a MLP $f_j(\cdot)$ to output the $Q$ value of the agent $j$, 
\begin{equation}
    Q_j (\mathbf{o}_j,a_j)=f_j (h_j (\mathbf{o}_j),\mathbf{x}_j )
\label{equ:Q_value}
\end{equation}
where $\mathbf{x}_j$ is the state of the remaining agents, with contributions to the action for agent $j$, $\mathbf{x}_j = \sum_{i \in \mathbf{Z}\backslash\{j\}}\alpha_i\mathbf{v}_i$, where the $\mathbf{Z}\backslash\{j\}$ denotes the set of z nearest agents to the agent $j$ among all agents, and $\alpha_i$ is the associated attention weight, which is defined as
\begin{equation}
    \alpha_i=\delta((\mathbf{q}_j \mathbf{k}_i^T)/\sqrt{d_k})
\label{equ:W_i}
\end{equation}
where $\delta(\cdot)$ is a softmax function, $d_k$ is the dimension of $\mathbf{k}_i$, $\mathbf{q}_j$ denotes state encoding of the agent $j$,
and $\mathbf{v}_i$ denotes the state-action encodings of the nearest z agents of all agents to agent $j$.
\section{Experiments}

\subsection{Experimental setup}
\begin{figure}[t]
\subfigure[Crowded environment \Rmnum{1}.]{
\includegraphics[width=0.24\textwidth]{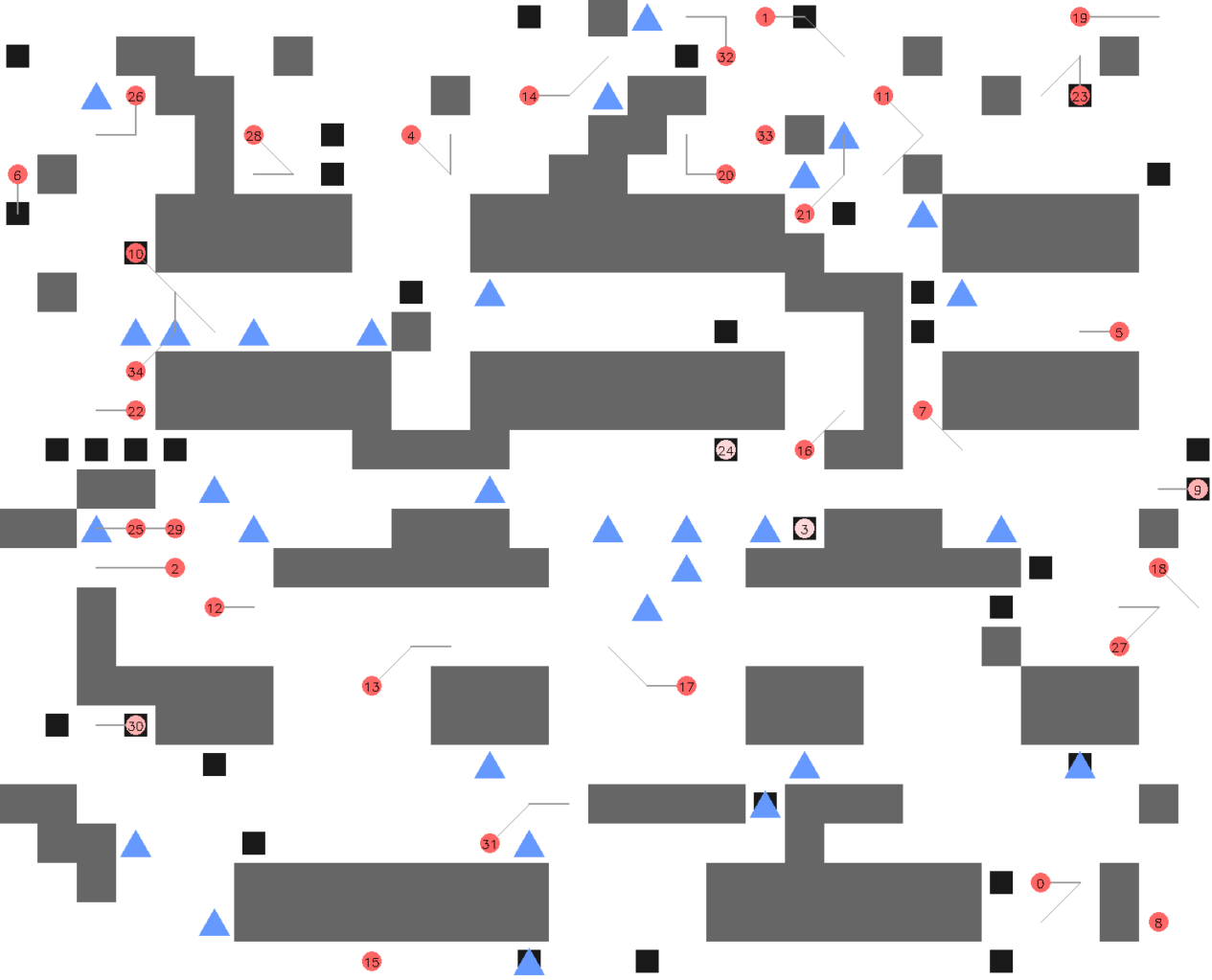}
}\hspace{-5mm}
\subfigure[Crowded environment \Rmnum{2}.]{
\includegraphics[width=0.195\textwidth]{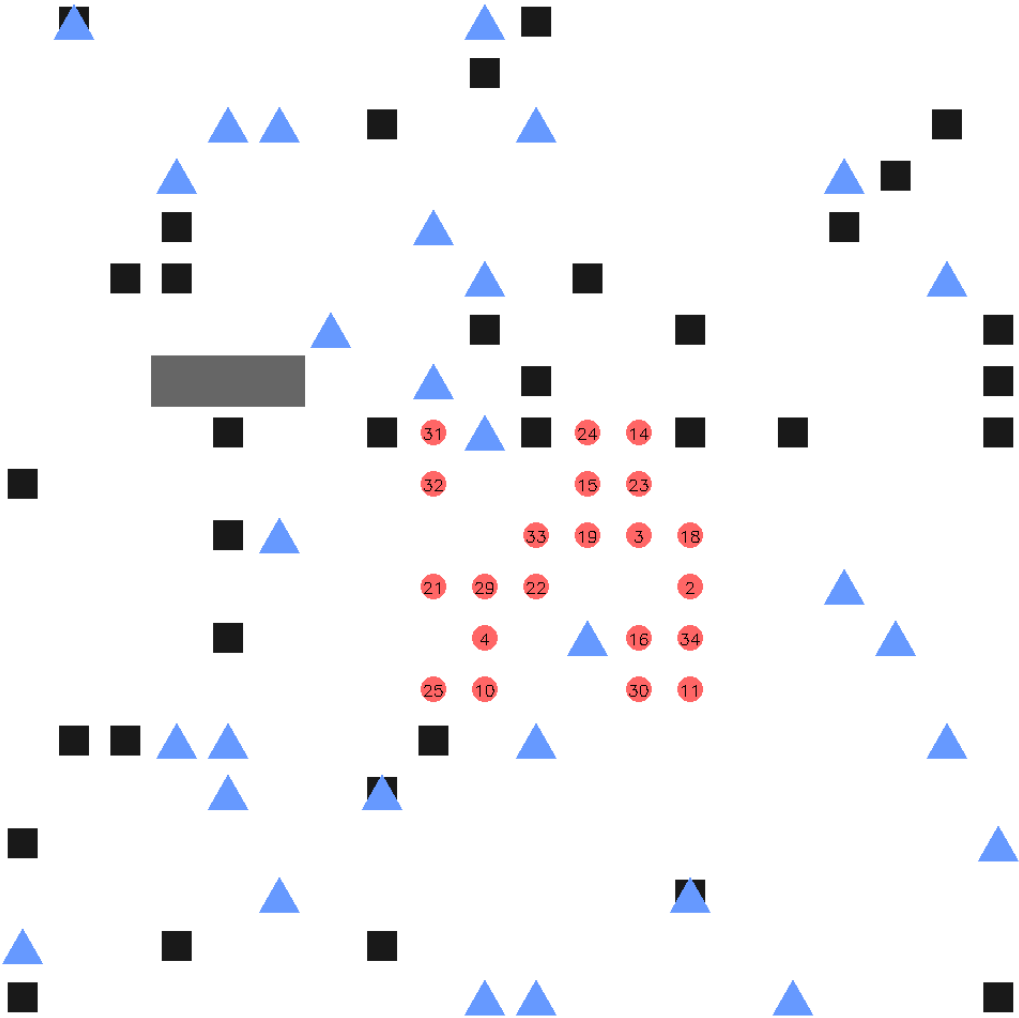}
}
\subfigure[Crowded environment \Rmnum{3}.]{
\includegraphics[width=0.225\textwidth]{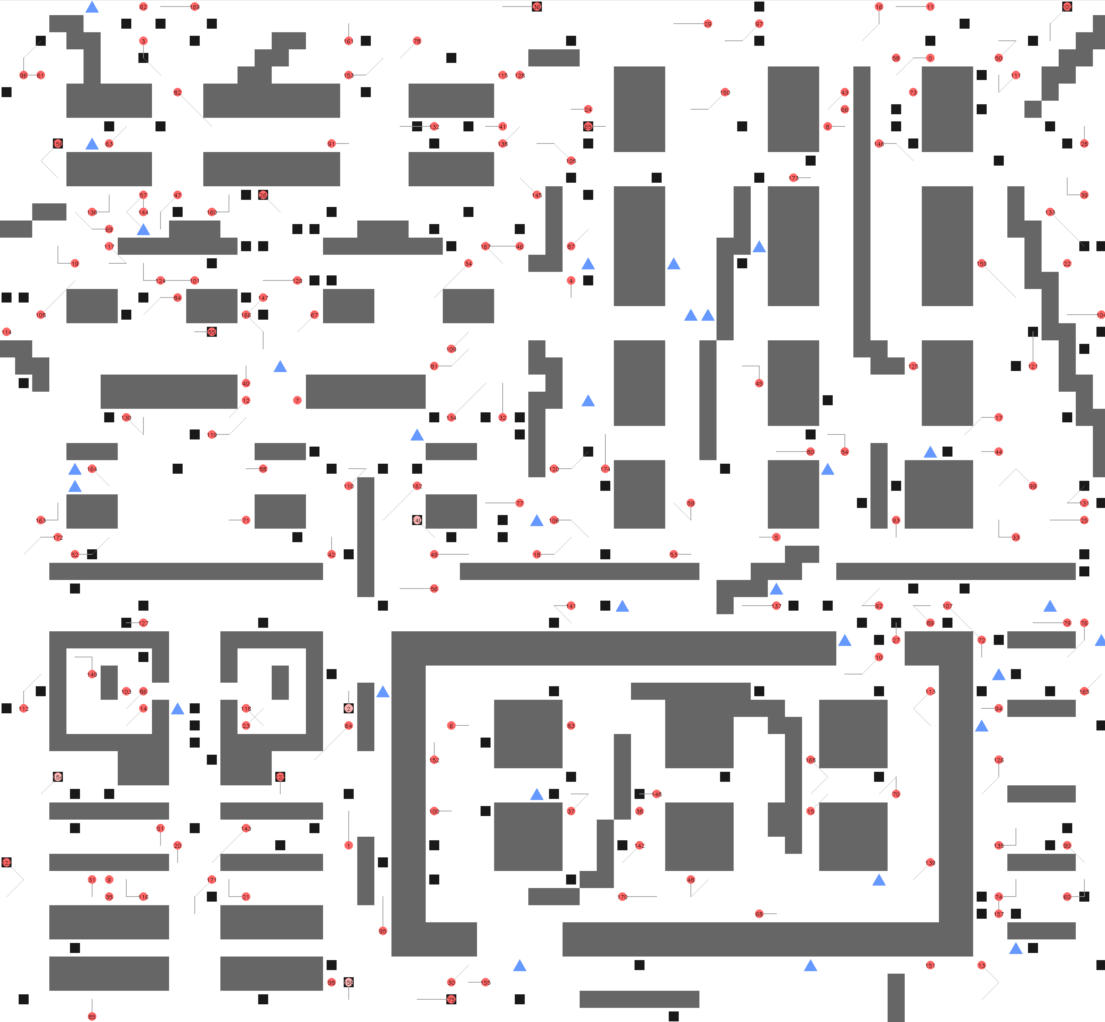}
}
\subfigure[Non-crowded environment.]{
\includegraphics[width=0.225\textwidth]{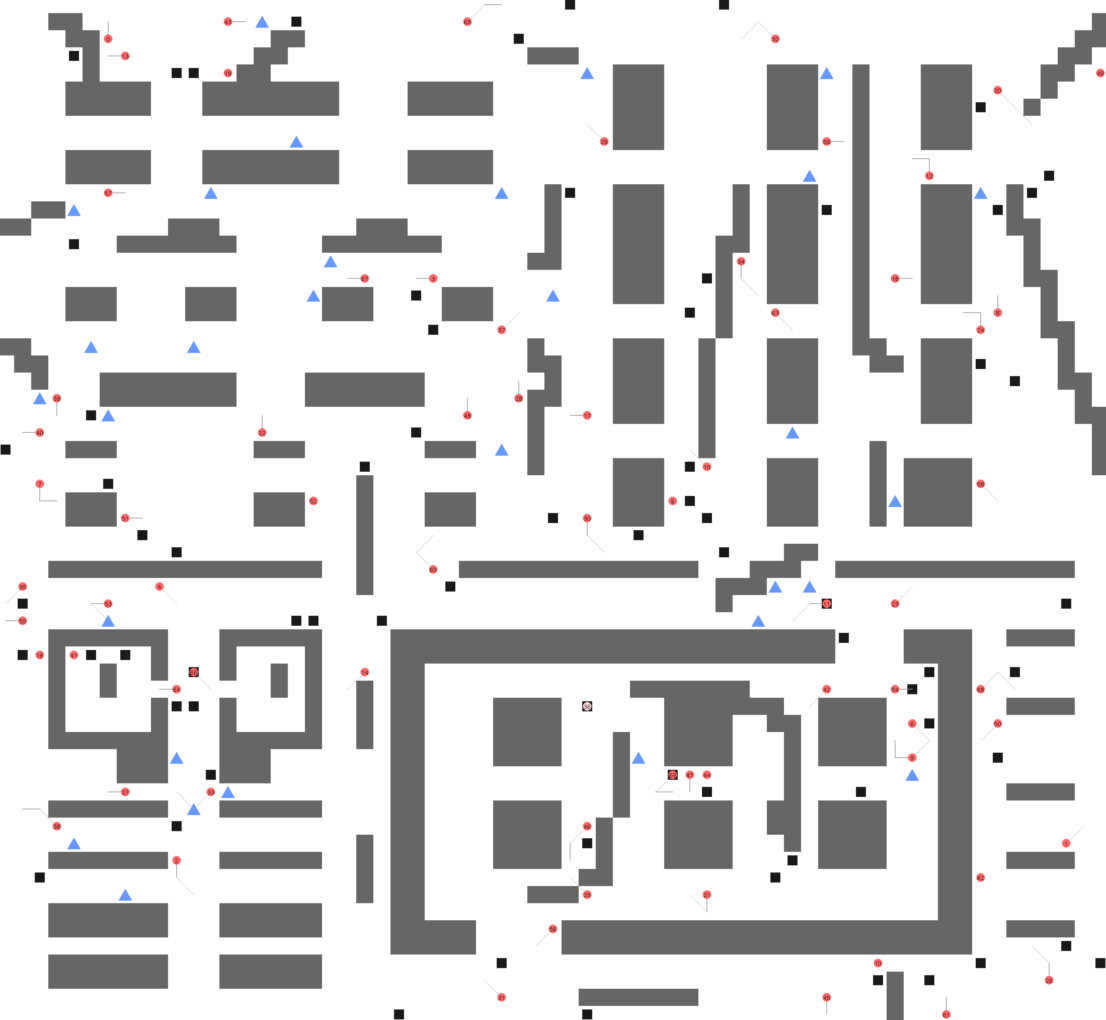}
}\hspace{-1mm}
\caption{The gray blocks are static obstacles, the orange circles represent the agents, the black blocks are the targets of the agents, and the blue triangles represent the dynamic obstacles. In every episode, the starting point is randomly initialized under the premise that the agent does not occupy the target point. In b, if one agent will occupy the position of another agent during the initialization, the agent waits for the other agent to move one step before initializing.}\label{fig:environment}
\vspace{-5mm}
\end{figure}

To fairly compare the performance of our AB-Mapper algorithm with the SOTA Mapper method,
we use the same grid world simulation environments.
Fig.~\ref{fig:environment} shows typical examples of the environments where the gray blocks are static obstacles, the orange circles represent the agents, the black blocks are the targets of the agents, and the blue triangles represent the dynamic obstacles. In these environments, each dynamic obstacle is navigated to a randomly selected target using the LRA* algorithm~\cite{silver2005cooperative}, and the agents must reach their respective target points within a prescribed maximum number of steps.  

In the MAPF test map set used by Li et al.~\cite{li2020graph}, the agent density is 0.025 and 0.05, respectively. We define the experimental environment with agent density greater than or equal to 0.025 as a crowded environment. In order to verify the ability of AB-Mapper to solve MAPF problems in a dynamic environment, especially in a dynamic crowded environment, we design dynamic crowded environments as (a), (b), and (c) and dynamic non-crowded environment as (d) in Fig.~\ref{fig:environment}. We take the success rate (the number of agents successfully reaching their targets over the total number of agents) as the evaluation index of the algorithm. In order to verify the improvement of communication and attention mechanism, we apply BicNet and attention mechanism to solve MAPF problems, respectively, and two ablation experiments (Mapper with only BicNet, Mapper with only attention). In the crowded environment \Rmnum{1}, \Rmnum{2} and \Rmnum{3} of Fig.~\ref{fig:environment}, the density of agents is about 0.044, 0.0875 and 0.044 (number of agents over environmental area), respectively. In the crowded environment \Rmnum{2} of Fig.~\ref{fig:environment}, we cluster the initialized positions of the smart bodies in the middle region to achieve the crowding effect. We also set up a non-crowded environment with an agent density of 0.017, as shown in Fig.~\ref{fig:environment}(d). In these four environments, the number of dynamic obstacles is 30. Because the MAPF test atlas has no dynamic obstacles, it is not comparable with our method.
\subsection{Experimental parameters}
In the experiments, we use the SOTA Mapper method as the BASELINE, and set the reward discount factor $\gamma = 0.99$ and the deviation route penalty weight $\lambda = 0.3$.
The parameters of Mapper are consistent with the original paper, the learning rate is 0.0003, and the evolutionary algorithm is used once every 100 episodes.
In addition to the basic experimental parameters,
in the ablation experiment of Mapper with only BicNet, the critic network and actor network are highly coupled, so there is only one learning rate, which is set to 0.00004.
In the ablation experiment of Mapper with only attention, the soft update parameter $\tau$ of the critic network is 0.001, and the learning rate of the critic network is 0.0001.
In our AB-Mapper network, the actor network has a learning rate of 0.00004, the critic network has a learning rate of 0.0001, and the critic network soft update parameter $\tau$ is 0.001.
In each class of environments, the number of neighbors that each agent pays attention to when using the selective local attention mechanism is shown in Table~\ref{table:params}, and we use distance as the selection criterion to choose only the state-action information of the nearest z agents for calculating the attention weights.
\begin{table}[h]
\vspace*{-3mm} 
\setlength{\abovecaptionskip}{-2mm}
\setlength{\belowcaptionskip}{-1mm}
\caption{Parameters setting}
\label{table_example}
\begin{center}
\begin{tabular}{|c|c|c|c|}
\hline Crowded environment &  Number of 
agents &  Dimension & z  \\
\hline \Rmnum{1} & 35  & $25 \times 31$ & 3 \\
\hline \Rmnum{2} & 35  & $20 \times 20$& 5 \\
\hline \Rmnum{3} & 175  & $60 \times 65$& 15 \\
\hline Non- & 70  & $60 \times 65$& 6 \\
\hline
\end{tabular}\label{table:params}
\end{center}
\vspace{-7mm}
\end{table}
\begin{figure*}[h]
\vspace*{8pt} 
\centering
\includegraphics[width=0.85\textwidth]{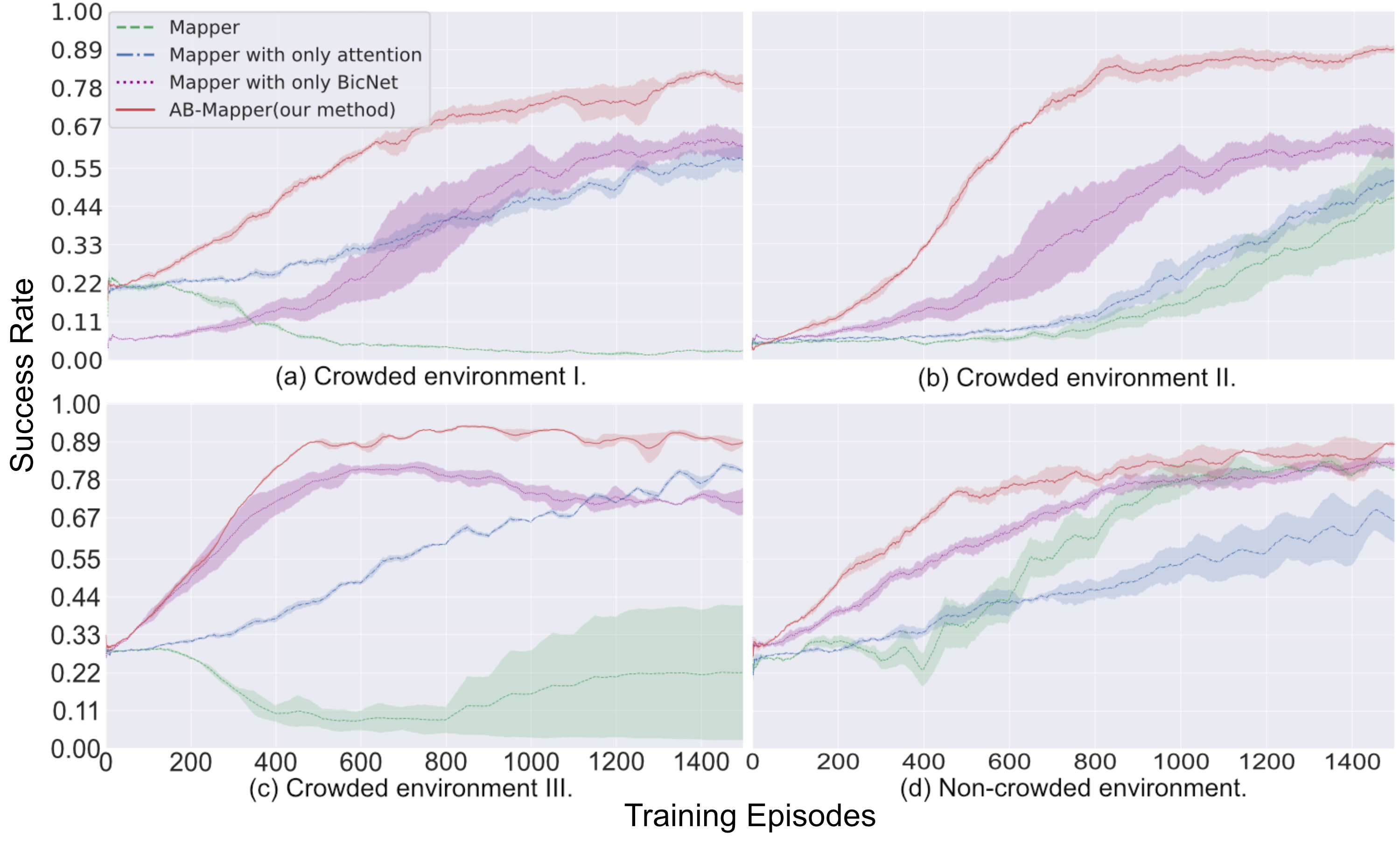}
\caption{The comparison result of success rate. It can be seen that when the environment becomes crowded, compared with AB-Mapper, Mapper has a low success rate and large variance.
AB-Mapper is the combination of the advantages of BicNet and attention mechanism in order to plan the path efficiently.}\label{fig:result}
\vspace{-4mm}
\end{figure*}
\subsection{Experimental results}
It can be seen in Fig.~\ref{fig:result} that when the environment becomes crowded, due to the interaction of agents and the interference of dynamic obstacles, the non-stationary of the environment increases, and Mapper cannot exchange information, so a coordinated strategy cannot be planned. In addition, Mapper's evolutionary algorithm will select a neural network model from the agent with the largest reward (the optimal agent) and give it to the agents with lower rewards, which is a risky practice. If the optimal agent fails to show good path planning ability in the initial training process, it is easy to lead to the failure of subsequent training or require a longer training time. This is the reason for the low success rate and large variance of Mapper in Fig.~\ref{fig:result}.

From the success rate curve of Mapper with only BicNet in Fig.~\ref{fig:result}, it can be seen that the success rate has grown faster before about the first 1000 episodes, which is because of the interaction of information can make the actor network plan a coordinated strategy.

After about 1000 episodes, the Mapper with only BicNet success rate curve has a slower growth rate, as shown in Fig.~\ref{fig:result}(a), (b) and (d). In particular, there has been a certain degree of decline in Fig.~\ref{fig:result}(c). The above phenomenon occurs because the interaction of information in BicNet is continuous, and the information of the first agent flows to the last agent, which cannot provide a single agent with accurate and highly relevant agent information. However, the curves of Mapper with only attention in all sub-graphs of Fig.~\ref{fig:result} can grow at a constant rate. In particular, the success rate of Mapper with only attention in Fig.~\ref{fig:result}(c) can exceed that of Mapper with only BicNet in the later stage of training, which is because the critic network with attention mechanism can accurately allocate attention weights to other agents that have a more significant impact, provide information about agents with solid relevance, and improve the stability of the policy during training. Finally, by comparing (c) and (d) in Fig.~\ref{fig:result}, the attention mechanism is also suitable for MAPF in the crowded environments. The final 
average success rate of each method could be observed from Table~\ref{table:comparison}.

By using the BicNet communication mechanism and selective local attention mechanism, AB-Mapper can reach the target point with a higher success rate than Mapper in dynamic environments, especially in crowded environments. This ensures the quality of path planning, and it is in favor of real applications.

\begin{table}[h]
\vspace{-3mm}
\setlength{\abovecaptionskip}{-5mm}
\setlength{\belowcaptionskip}{-1mm}
\caption{Comparison of success rate over different algorithm.}
\label{table_example}
\begin{center}
\begin{tabular}{|c|c|c|c|c|}
\hline \multicolumn{1}{|c|} {   } & \multicolumn{4}{c|} { Success Rate } \\
\hline \tabincell{c}{Crowded \\environment}  & Mapper & \tabincell{c}{Mapper with 
\\ only attention} & \tabincell{c}{Mapper with\\ only BicNet} & AB-Mapper \\
\hline \Rmnum{1} & 2.50\% &	57.06\% & 69.22\% & \bf{81.45\%} \\
\hline \Rmnum{2} & 45.87\% & 49.34\% & 62.36\% &	\bf{88.53\%} \\
\hline \Rmnum{3} & 21.86\% & 80.29\% & 72.36\% &	\bf{89.17\%} \\
\hline Non- & 81.56\% & 66.49\% & 82.53\% & \bf{85.86\%} \\
\hline
\end{tabular}\label{table:comparison}
\end{center}
\vspace{-5mm}
\end{table}
\section{Conclusion}
In order to enable efficient path planning for multi-agent in dynamic crowded environments, this paper proposes a MAPF method based on the actor-critic reinforcement learning framework. We mainly improve on the Mapper algorithm in two aspects: on the one hand, the actor network introduces BicNet as a communication module, which has a bidirectional recursive information transfer mechanism. Therefore the actor network can plan coordinated strategies through the interaction of information during the sequential decision-making process; on the other hand, we design a centralized critic network that can allocate attention weights to a limited number of other agents around each agent. In the path planning process, the critic network helps an agent learn how to accurately allocate attention weights to other agents that have a more significant impact on the current agent and effectively judge the agent’s policy. Experimental results show that AB-Mapper outperforms Mapper in terms of success rate in dynamic environments, ensuring the solution quality of MAPF problems, especially in dynamic crowded environments.

\section{Acknowledgement}
We thank Zuxin Liu for providing the Mapper source code. We would also like to thank Dr. Li Nan Nan from Macau University of Science and Technology for his valuable discussion on this paper.

\printbibliography

@inproceedings{vsvancara2019online,
  title={Online multi-agent pathfinding},
  author={{\v{S}}vancara, Jiri and Vlk, Marek and Stern, Roni and Atzmon, Dor and Bart{\'a}k, Roman},
  booktitle={Proceedings of the AAAI Conference on Artificial Intelligence},
  volume={33},
  number={01},
  pages={7732--7739},
  year={2019}
}

@inproceedings{stern2019multi,
  title={Multi-agent pathfinding: Definitions, variants, and benchmarks},
  author={Stern, Roni and Sturtevant, Nathan R and Felner, Ariel and Koenig, Sven and Ma, Hang and Walker, Thayne T and Li, Jiaoyang and Atzmon, Dor and Cohen, Liron and Kumar, TK Satish and others},
  booktitle={Twelfth Annual Symposium on Combinatorial Search},
  year={2019}
}

@article{lejeune2021survey,
  title={A Survey of the Multi-Agent Pathfinding Problem},
  author={Lejeune, Erwin and Sarkar, Sampreet and Jezequel, Lo{\i}g},
  year={2021}
}

@inproceedings{li2019symmetry,
  title={Symmetry-breaking constraints for grid-based multi-agent path finding},
  author={Li, Jiaoyang and Harabor, Daniel and Stuckey, Peter J and Ma, Hang and Koenig, Sven},
  booktitle={Proceedings of the AAAI Conference on Artificial Intelligence},
  volume={33},
  number={01},
  pages={6087--6095},
  year={2019}
}

@inproceedings{felner2017search,
  title={Search-based optimal solvers for the multi-agent pathfinding problem: Summary and challenges},
  author={Felner, Ariel and Stern, Roni and Shimony, Solomon Eyal and Boyarski, Eli and Goldenberg, Meir and Sharon, Guni and Sturtevant, Nathan and Wagner, Glenn and Surynek, Pavel},
  booktitle={Tenth Annual Symposium on Combinatorial Search},
  year={2017}
}

@article{sharon2015conflict,
  title={Conflict-based search for optimal multi-agent pathfinding},
  author={Sharon, Guni and Stern, Roni and Felner, Ariel and Sturtevant, Nathan R},
  journal={Artificial Intelligence},
  volume={219},
  pages={40--66},
  year={2015},
  publisher={Elsevier}
}

@article{silver2005cooperative,
  title={Cooperative Pathfinding.},
  author={Silver, David},
  journal={Aiide},
  volume={1},
  pages={117--122},
  year={2005},
  publisher={Marina Del Rey}
}

@article{hasan2020defensive,
  title={Defensive Escort Teams for Navigation in Crowds via Multi-Agent Deep Reinforcement Learning},
  author={Hasan, Yazied A and Garg, Arpit and Sugaya, Satomi and Tapia, Lydia},
  journal={IEEE Robotics and Automation Letters},
  volume={5},
  number={4},
  pages={5645--5652},
  year={2020},
  publisher={IEEE}
}

@inproceedings{liu2020mapper,
  title={Mapper: Multi-agent path planning with evolutionary reinforcement learning in mixed dynamic environments},
  author={Liu, Zuxin and Chen, Baiming and Zhou, Hongyi and Koushik, Guru and Hebert, Martial and Zhao, Ding},
  booktitle={2020 IEEE/RSJ International Conference on Intelligent Robots and Systems (IROS)},
  pages={11748--11754},
  year={2020},
  organization={IEEE}
}

@inproceedings{chen2017decentralized,
  title={Decentralized non-communicating multiagent collision avoidance with deep reinforcement learning},
  author={Chen, Yu Fan and Liu, Miao and Everett, Michael and How, Jonathan P},
  booktitle={2017 IEEE international conference on robotics and automation (ICRA)},
  pages={285--292},
  year={2017},
  organization={IEEE}
}

@article{lyu2021contrasting,
  title={Contrasting centralized and decentralized critics in multi-agent reinforcement learning},
  author={Lyu, Xueguang and Xiao, Yuchen and Daley, Brett and Amato, Christopher},
  journal={arXiv preprint arXiv:2102.04402},
  year={2021}
}

@inproceedings{omidshafiei2017deep,
  title={Deep decentralized multi-task multi-agent reinforcement learning under partial observability},
  author={Omidshafiei, Shayegan and Pazis, Jason and Amato, Christopher and How, Jonathan P and Vian, John},
  booktitle={International Conference on Machine Learning},
  pages={2681--2690},
  year={2017},
  organization={PMLR}
}

@inproceedings{son2019qtran,
  title={Qtran: Learning to factorize with transformation for cooperative multi-agent reinforcement learning},
  author={Son, Kyunghwan and Kim, Daewoo and Kang, Wan Ju and Hostallero, David Earl and Yi, Yung},
  booktitle={International Conference on Machine Learning},
  pages={5887--5896},
  year={2019},
  organization={PMLR}
}

@inproceedings{wadhwania2019policy,
  title={Policy distillation and value matching in multiagent reinforcement learning},
  author={Wadhwania, Samir and Kim, Dong-Ki and Omidshafiei, Shayegan and How, Jonathan P},
  booktitle={2019 IEEE/RSJ International Conference on Intelligent Robots and Systems (IROS)},
  pages={8193--8200},
  year={2019},
  organization={IEEE}
}

@inproceedings{choi2020fast,
  title={Fast adaptation of deep reinforcement learning-based navigation skills to human preference},
  author={Choi, Jinyoung and Dance, Christopher and Kim, Jung-eun and Park, Kyung-sik and Han, Jaehun and Seo, Joonho and Kim, Minsu},
  booktitle={2020 IEEE International Conference on Robotics and Automation (ICRA)},
  pages={3363--3370},
  year={2020},
  organization={IEEE}
}

@inproceedings{wen2020smix,
  title={Smix ($\lambda$): Enhancing centralized value functions for cooperative multi-agent reinforcement learning},
  author={Wen, Chao and Yao, Xinghu and Wang, Yuhui and Tan, Xiaoyang},
  booktitle={Proceedings of the AAAI Conference on Artificial Intelligence},
  volume={34},
  number={05},
  pages={7301--7308},
  year={2020}
}

@article{sukhbaatar2016learning,
  title={Learning multiagent communication with backpropagation},
  author={Sukhbaatar, Sainbayar and Fergus, Rob and others},
  journal={Advances in neural information processing systems},
  volume={29},
  pages={2244--2252},
  year={2016}
}

@article{singh2018learning,
  title={Learning when to communicate at scale in multiagent cooperative and competitive tasks},
  author={Singh, Amanpreet and Jain, Tushar and Sukhbaatar, Sainbayar},
  journal={arXiv preprint arXiv:1812.09755},
  year={2018}
}

@article{foerster2016learning,
  title={Learning to communicate with deep multi-agent reinforcement learning},
  author={Foerster, Jakob N and Assael, Yannis M and De Freitas, Nando and Whiteson, Shimon},
  journal={arXiv preprint arXiv:1605.06676},
  year={2016}
}

@article{wang2020mobile,
  title={Mobile robot path planning in dynamic environments through globally guided reinforcement learning},
  author={Wang, Binyu and Liu, Zhe and Li, Qingbiao and Prorok, Amanda},
  journal={IEEE Robotics and Automation Letters},
  volume={5},
  number={4},
  pages={6932--6939},
  year={2020},
  publisher={IEEE}
}

@article{riviere2020glas,
  title={Glas: Global-to-local safe autonomy synthesis for multi-robot motion planning with end-to-end learning},
  author={Riviere, Benjamin and H{\"o}nig, Wolfgang and Yue, Yisong and Chung, Soon-Jo},
  journal={IEEE Robotics and Automation Letters},
  volume={5},
  number={3},
  pages={4249--4256},
  year={2020},
  publisher={IEEE}
}

@article{sartoretti2019primal,
  title={Primal: Pathfinding via reinforcement and imitation multi-agent learning},
  author={Sartoretti, Guillaume and Kerr, Justin and Shi, Yunfei and Wagner, Glenn and Kumar, TK Satish and Koenig, Sven and Choset, Howie},
  journal={IEEE Robotics and Automation Letters},
  volume={4},
  number={3},
  pages={2378--2385},
  year={2019},
  publisher={IEEE}
}

@article{jiang2018graph,
  title={Graph convolutional reinforcement learning},
  author={Jiang, Jiechuan and Dun, Chen and Huang, Tiejun and Lu, Zongqing},
  journal={arXiv preprint arXiv:1810.09202},
  year={2018}
}

@article{ma2021distributed,
  title={Distributed Heuristic Multi-Agent Path Finding with Communication},
  author={Ma, Ziyuan and Luo, Yudong and Ma, Hang},
  journal={arXiv preprint arXiv:2106.11365},
  year={2021}
}

@article{zhang2020robot,
  title={Robot Navigation among External Autonomous Agents through Deep Reinforcement Learning using Graph Attention Network},
  author={Zhang, Tianle and Qiu, Tenghai and Pu, Zhiqiang and Liu, Zhen and Yi, Jianqiang},
  journal={IFAC-PapersOnLine},
  volume={53},
  number={2},
  pages={9465--9470},
  year={2020},
  publisher={Elsevier}
}

@article{zhou2021robot,
  title={Robot Navigation in a Crowd by Integrating Deep Reinforcement Learning and Online Planning},
  author={Zhou, Zhiqian and Zhu, Pengming and Zeng, Zhiwen and Xiao, Junhao and Lu, Huimin and Zhou, Zongtan},
  journal={arXiv preprint arXiv:2102.13265},
  year={2021}
}

@article{chen2020robot,
  title={Robot navigation in crowds by graph convolutional networks with attention learned from human gaze},
  author={Chen, Yuying and Liu, Congcong and Shi, Bertram E and Liu, Ming},
  journal={IEEE Robotics and Automation Letters},
  volume={5},
  number={2},
  pages={2754--2761},
  year={2020},
  publisher={IEEE}
}

@inproceedings{rosbach2020planning,
  title={Planning on the fast lane: Learning to interact using attention mechanisms in path integral inverse reinforcement learning},
  author={Rosbach, Sascha and Li, Xing and Gro{\ss}johann, Simon and Homoceanu, Silviu and Roth, Stefan},
  booktitle={2020 IEEE/RSJ International Conference on Intelligent Robots and Systems (IROS)},
  pages={5187--5193},
  year={2020},
  organization={IEEE}
}

@article{shah2018follownet,
  title={Follownet: Robot navigation by following natural language directions with deep reinforcement learning},
  author={Shah, Pararth and Fiser, Marek and Faust, Aleksandra and Kew, J Chase and Hakkani-Tur, Dilek},
  journal={arXiv preprint arXiv:1805.06150},
  year={2018}
}

@inproceedings{chen2019crowd,
  title={Crowd-robot interaction: Crowd-aware robot navigation with attention-based deep reinforcement learning},
  author={Chen, Changan and Liu, Yuejiang and Kreiss, Sven and Alahi, Alexandre},
  booktitle={2019 International Conference on Robotics and Automation (ICRA)},
  pages={6015--6022},
  year={2019},
  organization={IEEE}
}

@article{peng2017multiagent,
  title={Multiagent bidirectionally-coordinated nets: Emergence of human-level coordination in learning to play starcraft combat games},
  author={Peng, Peng and Wen, Ying and Yang, Yaodong and Yuan, Quan and Tang, Zhenkun and Long, Haitao and Wang, Jun},
  journal={arXiv preprint arXiv:1703.10069},
  year={2017}
}

@inproceedings{iqbal2019actor,
  title={Actor-attention-critic for multi-agent reinforcement learning},
  author={Iqbal, Shariq and Sha, Fei},
  booktitle={International Conference on Machine Learning},
  pages={2961--2970},
  year={2019},
  organization={PMLR}
}

@article{parnika2021attention,
  title={Attention Actor-Critic algorithm for Multi-Agent Constrained Co-operative Reinforcement Learning},
  author={Parnika, P and Diddigi, Raghuram Bharadwaj and Danda, Sai Koti Reddy and Bhatnagar, Shalabh},
  journal={arXiv preprint arXiv:2101.02349},
  year={2021}
}

@inproceedings{li2020graph,
  title={Graph neural networks for decentralized multi-robot path planning},
  author={Li, Qingbiao and Gama, Fernando and Ribeiro, Alejandro and Prorok, Amanda},
  booktitle={2020 IEEE/RSJ International Conference on Intelligent Robots and Systems (IROS)},
  pages={11785--11792},
  year={2020},
  organization={IEEE}
}

@inproceedings{vaswani2017attention,
  title={Attention is all you need},
  author={Vaswani, Ashish and Shazeer, Noam and Parmar, Niki and Uszkoreit, Jakob and Jones, Llion and Gomez, Aidan N and Kaiser, {\L}ukasz and Polosukhin, Illia},
  booktitle={Advances in neural information processing systems},
  pages={5998--6008},
  year={2017}
}
\end{document}